\pdfoutput=1

\documentclass[11pt]{article}

\usepackage[]{emnlp21/emnlp2021}

\usepackage{times}
\usepackage{latexsym}
\usepackage{verbatim}
\usepackage{booktabs}
\usepackage{subcaption}
\usepackage{graphicx}
\usepackage{xcolor,colortbl}

\usepackage[T1]{fontenc}

\usepackage[utf8]{inputenc}

\usepackage{microtype}

\newcommand{\mysubsection}[1]{\vspace{0.3em} \noindent\textbf{#1}}

\title{Analyzing the Surprising Variability in Word Embedding Stability\\ Across Languages}

\author{Laura Burdick, Jonathan K. Kummerfeld \and Rada Mihalcea \\
  Computer Science \& Engineering \\
  University of Michigan, Ann Arbor \\
  \texttt{\{lburdick,jkummerf,mihalcea\}@umich.edu}}

\date{}

\begin{document}
\maketitle
\begin{abstract}
Word embeddings are powerful representations that form the foundation of many natural language processing architectures, both in English and in other languages.
To gain further insight into word embeddings, we explore their stability (e.g., overlap between the nearest neighbors of a word in different embedding spaces) in diverse languages.
We discuss linguistic properties that are related to stability, drawing out insights about correlations with affixing, language gender systems, and other features. 
This has implications for embedding use, particularly in research that uses them to study language trends.
\end{abstract}

\section{Introduction}
Word embeddings have become an established part of natural language processing (NLP) \cite{collobert2011natural,wang2020survey}. Stability, defined as the overlap between the nearest neighbors of a word in different embedding spaces, was introduced to measure variations in local embedding neighborhoods across changes in data, algorithms, and word properties \cite{antoniak2018evaluating,Wendlandt18Surprising}. These studies found that many common English embedding spaces are surprisingly unstable, which has implications for work that uses embeddings as features in downstream tasks, and work that uses embeddings to study specific properties of language.

However, research to date on word embedding stability has been exclusively done on English and so is not representative of all languages.
In this work, we explore the stability of word embeddings in a wide range of languages. Better understanding the differences caused by diverse languages will provide a foundation for building embeddings and NLP tools in all languages.\footnote{
Code is available at \url{https://lit.eecs.umich.edu/downloads.html}.
}

In English and other very high resource languages, it has become common practice to use contextualized word embeddings, such as BERT \cite{devlin2018bert} and XLNet \cite{yang2019xlnet}. These algorithms require huge amounts of computational resources and data. For example, it takes 2.5 days to train XLNet with 512 TPU v3 chips. In addition to requiring heavy computational resources, most contextualized embedding algorithms need large amounts of data. BERT uses 3.3 billion words of training data. In contrast to these large corpora, many datasets from low-resource languages are fairly small \cite{maxwell2006frontiers}.
To support scenarios where using huge amounts of data and computational resources is not feasible, it is important to continue developing our understanding of context-independent word embeddings, such as word2vec \cite{mikolov2013distributed} and GloVe \cite{pennington2014glove}. These algorithms continue to be used in a wide variety of situations, including the computational humanities \cite{abdulrahimideological,hellrich2019modeling} and languages where only small corpora are available \cite{joshi2019word}. 

In this work, we consider how stability varies for different languages, and how linguistic properties are related to stability---a previously understudied relationship.
Using regression modeling, we capture relationships between linguistic properties and average stability of a language, and we draw out insights about how linguistic features relate to stability. For instance, we find that embeddings in languages with more affixing tend to be less stable.
Our findings provide crucial context for research that uses word embeddings to study language properties and trends \cite[e.g.,][]{heyman2019can,abdulrahimideological}, which often rely on raw embeddings created by GloVe or word2vec. If these embeddings are unstable, then research using them needs to take this into account in terms of methodologies and error analysis.

\section{Related Work}

Word embeddings are low-dimensional vectors used to represent words, normally in downstream tasks, such as word sense disambiguation \cite{scarlini2020more} and text summarization \cite{moradi2020summarization}. They have been shown to capture both syntactic and semantic properties of words, making them useful in a wide range of NLP tasks \cite{wang2020static}. In this work, we explore word embeddings that generate one embedding per word, regardless of the word's context.
We consider two widely used algorithms: word2vec \cite{mikolov2013distributed} and GloVe \cite{pennington2014glove}.

Our work analyzes embeddings in multiple languages, which is important because embeddings are commonly used across many languages.
In particular, there has been interest in embeddings for low-resource languages \cite{chimalamarri2020morphological,stringham-izbicki-2020-evaluating}.

In this work, we use stability to measure the quality of word embeddings.
Similar to the work we present here on stability, other research looks at how nearest neighbors vary as properties of the embedding spaces change. \citet{pierrejean-tanguy-2018-towards} found that the lowest frequency and the highest frequency words have the highest variation among nearest neighbors.  Additional research has explored how semantic and syntactic properties of words change with different embedding algorithm and parameter choices \cite{artetxe-etal-2018-uncovering,yaghoobzadeh-schutze-2016-intrinsic}.
Unlike our work, previous studies only considered English.

Finally, while our work is \emph{not} a form of embedding evaluation, it is related to the topic \cite{chiu-etal-2016-intrinsic,rogers-etal-2018-whats,qiu2018revisiting}.
There has been extensive work on evaluating word embeddings, seen in the recent RepEval workshops \cite{ws-2019-evaluating}, and going back to work comparing them with counting based methods \cite{baroni-etal-2014-dont}.
Our findings indicate that work on embedding evaluation should take into consideration stability, using multiple training runs to confirm results.
Similarly, stability should be considered when studying the impact of embeddings on downstream tasks. \citet{MLSYS2020_c9e1074f} specifically looked at the downstream instability of word embeddings, and found that there is a stability-memory tradeoff, and higher stability can be achieved by increasing the embedding dimension. 

\section{Data}

In order to explore the stability of word embeddings in different languages,
we work with two datasets, Wikipedia and the Bible. While Wikipedia has more data, the Bible covers more languages. Wikipedia is a comparable corpus, whereas the Bible is a parallel corpus.

\mysubsection{Wikipedia Corpus.}
We use pre-processed Wikipedia dumps in 40 languages taken from \citet{al2013polyglot}.\footnote{Available online at \url{https://sites.google.com/site/rmyeid/projects/polyglot}.}
The size of these Wikipedia corpora varies from 329,136 sentences (Tagalog) to 75,241,648 sentences (English), with an average of 9,292,394 sentences. For all of our experiments, we downsample each corpus to work with comparably sized data (details in Section \ref{sec:stab_wiki}).

\mysubsection{Bible Corpus.}
We consider 97 languages from the pre-processed Bible corpus \cite{mccarthy-etal-2020-johns}:\footnote{Available by contacting \citet{mccarthy-etal-2020-johns}.} all languages for which at least 75\% of the Bible ($\geq23,326$ verses) is present.\footnote{To work with a maximum number of languages, we only consider the complete Protestant Bible (i.e., all of the verses that appear in the English King James Version of the Bible).}
This excludes many languages for which there is only a partial Bible, e.g., just the New Testament, which would be insufficient for training word vectors.
We consider two sets of languages with the Bible corpus: languages that overlap with the set of Wikipedia languages (26 languages), and all languages in the Bible corpus (97 languages).

\mysubsection{WALS.}
To gain linguistic properties of these languages, we use the World Atlas of Language Structures (WALS),\footnote{Available online at \url{https://wals.info}.} a database of phonological, lexical, and grammatical properties for over 2,000 languages \cite{wals}. This expert-curated resource contains 192 language features.
For example, WALS records subject, object, and verb word order for various languages.

\section{Calculating Stability in Many Languages}\label{sec:stability}

The first part of our work is a comparison of stability across languages.
Before presenting our measurements, we define stability and analyze some important methodological decisions.

\subsection{Defining Stability} \label{sec:background}

Stability is defined as the percent overlap between nearest neighbors in an embedding space. To calculate stability, given a word $W$ and two embedding spaces $A$ and $B$, take the ten nearest neighbors (measured using cosine similarity) of $W$ in both $A$ and $B$. The stability of $W$ is the percent overlap between these two lists of nearest neighbors.\footnote{While alternative definitions of stability are possible, e.g., considering a vector of similarities with a large set of words, we chose to use a prior definition of stability that has been rigorously studied. Similarly, sets of nearest neighbors smaller and larger than ten have been tried previously, with comparable results \cite{Wendlandt18Surprising}.} 100\% stability indicates perfect agreement between the two embedding spaces, while 0\% stability indicates complete disagreement. Table ~\ref{tab:stability_example} shows a simple example. This definition of stability can be generalized to more than two embedding spaces by considering the average overlap between pairs of embedding spaces. Let $X$ and $Y$ be two sets of embedding spaces. Then, for every pair of embedding spaces $(x, y)$, where $x \in X$ and $y \in Y$, take the ten nearest neighbors of $W$ in both $x$ and $y$ and calculate percent overlap. Let the stability be the average percent overlap over every pair of embedding spaces $(x, y)$.

\begin{table}[!tb]
\small
    \centering
    \begin{tabular}{c}
        \toprule
        Model 1: \textbf{indie}, \textbf{punk}, \textbf{progressive}, \textbf{pop}, \textit{roll}, \textbf{band}, \\blues, brass, class, \textbf{alternative} \\
        \midrule
        Model 2: \textbf{punk}, \textbf{indie}, \textbf{alternative}, \textbf{progressive}, \textbf{band}, \\ sedimentary, bands, psychedelic, \textit{climbing}, \textbf{pop} \\
        \midrule
        Model 3: \textbf{punk}, \textbf{pop}, \textbf{indie}, \textbf{alternative}, \textbf{band}, \textit{roll}, \\ \textbf{progressive}, folk, \textit{climbing}, metal \\
         \bottomrule
    \end{tabular}
    \caption{\label{tab:stability_example}
    Ten nearest neighbors for the word \textbf{rock} in three GloVe models trained on different subsets of Large English Wikipedia. Words in all lists are in bold; words in only two lists are italicized. Models 1 and 2 have 6 words (60\%) in common, models 1 and 3 have 7, and models 2 and 3 have 7.
    Therefore, this word has a stability of 66.7\%, the average word overlap between the three models.}
\end{table}

Previous work has explored stability for English word embeddings.
For instance, it was found that the presence of certain documents in the training corpus affects stability \cite{antoniak2018evaluating}, and that training and evaluating embeddings on separate domains is less stable than training and evaluating on the same domain \cite{Wendlandt18Surprising}.
In this work, we expand this analysis to a more diverse set of languages.

\subsubsection{The Effect of Downsampling on Stability} \label{sec:sampling}

Stability measures how changes to the input data or training algorithm affect the resulting embeddings.
Sometimes we make changes with the goal of shifting the embeddings, such as increasing the context window size to try to get embeddings that capture semantics more than syntax.
In other cases, we would hope a change would not substantially change embeddings, such as changing the random seed for the algorithm.
For our experiments, we consider a previously unstudied source of instability: different data samples from the same distribution.
This is a case where we hope embeddings remain stable, given a sufficiently large sample.

We generate data samples by downsampling a corpus to create multiple smaller corpora; we then measure stability across these downsamples.
The choice of sampling with or without replacement, and the size of the sample are subtle methodological choices.
In this section, we consider whether stability across downsamples produces consistent results that we can compare across languages.

\begin{figure}[!tb]
    \centering
    \begin{subfigure}[b]{0.45\textwidth}
        \includegraphics[width=\textwidth]{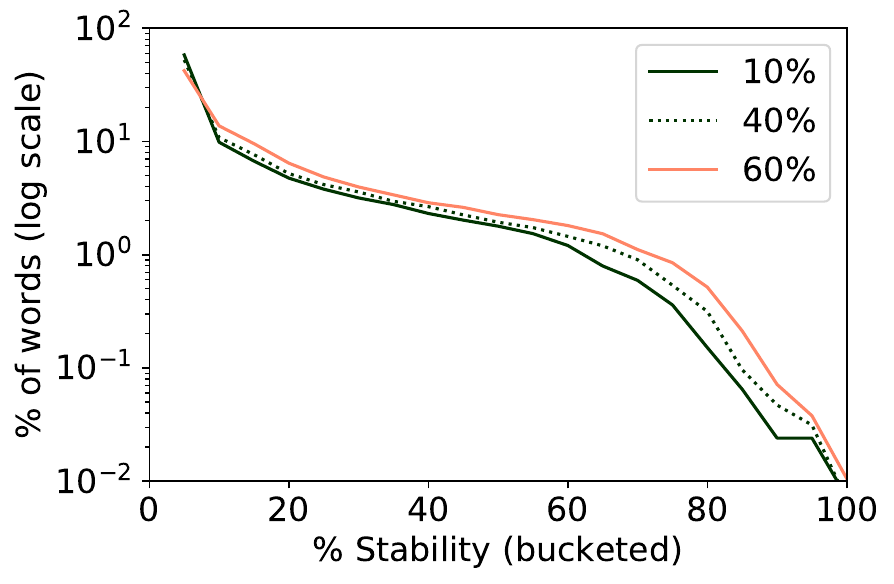}
        \caption{\label{fig:wikipedia_with_replacement}
        Sampling \emph{with} replacement, varying percentage overlap between samples.
        }
    \end{subfigure}

    \begin{subfigure}[b]{0.45\textwidth}
        \includegraphics[width=\textwidth]{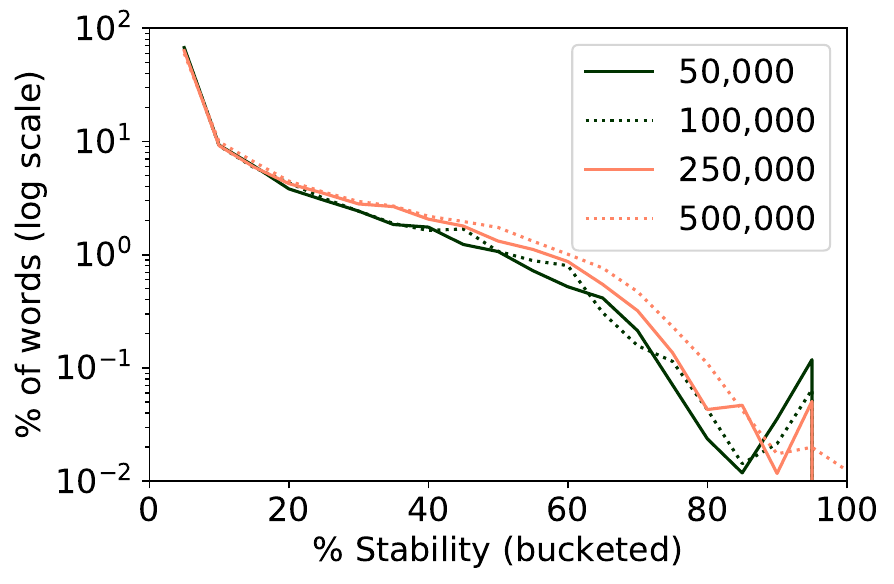}
        \caption{\label{fig:wikipedia_without_replacement}
        Sampling \emph{without} replacement, varying sample size.
        }
    \end{subfigure}
    \caption{
    Measuring the impact of data sampling parameters on stability measurements.
    Results when sampling \emph{with} replacement consistently increase as overlap increases (a).
    This poses a problem, as results may reflect corpus size rather than intrinsic stability.
    Results when sampling \emph{without} replacement do show a consistent pattern, even when the sample is only 50,000 sentences, a tenth of the largest sample size (b).
    }
\end{figure}

\begin{table*}[!t]
    \centering
    \small
    \begin{tabular}{lll}
    \toprule
    Experiment & Machine & Timing \\
    \midrule
    Training one w2v embedding on one Wikipedia corpus (Section~\ref{sec:stability}) & Machine 1 & 13 sec. \\
    Training one GloVe embedding on one Wikipedia corpus (Section~\ref{sec:stability}) & Machine 1 & 12 min. \\
    Calculating stability on one Wikipedia corpus (Section~\ref{sec:stability}) & Machine 1 & 17 sec. \\
    Training one w2v embedding on one Bible corpus (Section~\ref{sec:stability}) & Machine 1 & 5 sec. \\
    Calculating stability on one Bible corpus (Section~\ref{sec:stability}) & Machine 1 & 12 sec. \\
    Training regression model (Section~\ref{sec:regression}) & Machine 2 & < 7 sec.\\
    Leave-one-out cross-validation (Section~\ref{sec:results}) & Machine 2 & < 4 sec.\\
    \bottomrule
    \end{tabular}
    \caption{Runtimes for different experimental portions of this work. Machine 1 is four Intel(R) Xeon(R) CPU E5-1603 v3 @ 2.80 GHz processors. Machine 2 is a 2.9GHz Dual-Core Intel Core i5.}
    \label{tab:runtimes}
\end{table*}

First, we consider downsampling with replacement, shown in Figure~\ref{fig:wikipedia_with_replacement}. We use data drawn from an English Wikipedia corpus of 5,269,686 sentences (denoted ``Large English Wikipedia").\footnote{This data was used in \citet{tsvetkov-etal-2016-learning} and is available by contacting the authors of that paper.}
We randomly sample five sets of 500,000 sentences multiple times, controlling the amount of overlap between downsamples (from 10\% to 60\% shared across all five samples). For a specific overlap amount X\%, X\% of 500,000 sentences is randomly sampled and included in all of the five downsamples. The remaining (100-X)\% sentences are randomly sampled for each downsample.

Stability is calculated using GloVe embeddings and the words that occur in every downsample for every overlap percentage.
In Figure~\ref{fig:wikipedia_with_replacement}, we group stability into buckets of size 5\% (i.e., 0-5\%, 5-10\%, etc).
This allows us to see patterns in stability that are not visible from a single statistic, such as the overall average.
We see that while stability trends are similar for different overlap amounts, stability is consistently higher as the overlap amount increases.
This means that if we use downsampling with replacement, we cannot reliably compare stability across multiple corpora of varying sizes (e.g., Wikipedia and the much smaller Bible corpus). The overlap amount would change depending on the size of the corpus, changing our stability measurement.

Instead of downsampling with replacement, we consider downsampling without replacement, shown in Figure~\ref{fig:wikipedia_without_replacement} for different downsample sizes.
We see that varying the size of the downsample does not have a large effect on the patterns of stability. Particularly when looking at lower stability, the trends are remarkably consistent, even when the downsample size varies from 50,000 sentences to 500,000 sentences. The pattern grows less consistent when looking at higher stability, especially with smaller downsample sizes.

This comparison (Figures \ref{fig:wikipedia_with_replacement} and \ref{fig:wikipedia_without_replacement}) shows that downsampling without replacement produces more consistent (and thus comparable) stability results than downsampling with replacement.
Thus, we only consider downsampling without replacement.

\subsection{Stability for Wikipedia and the Bible} \label{sec:stab_wiki}

\begin{figure*}[!htb]
    \centering
    \includegraphics[width=\textwidth]{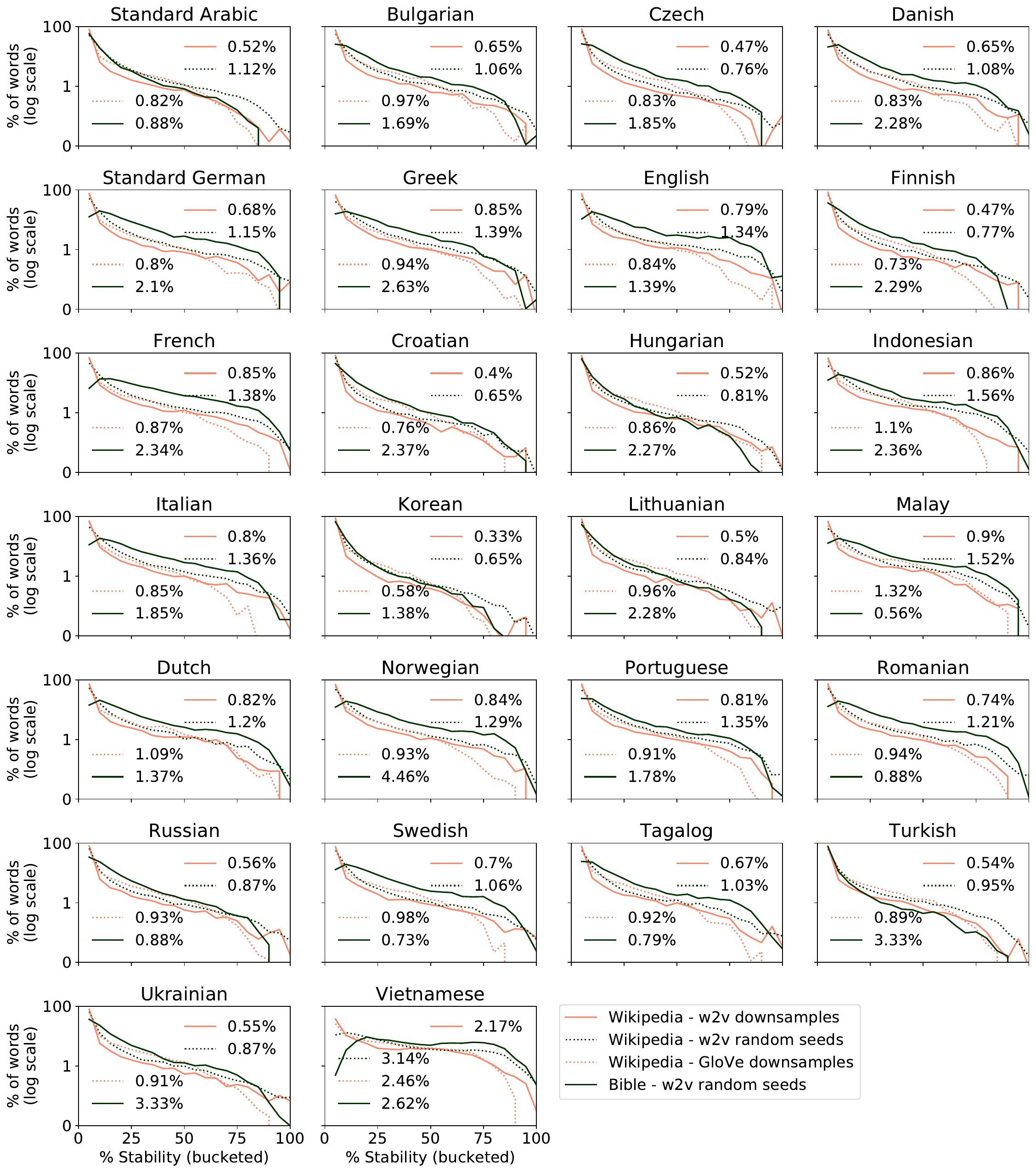}
    \caption{Percentage of words that occur in each stability bucket for four different methods, three on Wikipedia and one on the Bible. The 26 languages in common are shown here. The average stability for each method is shown on the individual graphs.}
    \label{fig:polyglot_methods}
\end{figure*}

Our first study, shown in Figure~\ref{fig:polyglot_methods}, considers stability across the 26 languages included in both Wikipedia and the Bible.
These results show three settings for Wikipedia:
(1) Stability of GloVe embeddings across five downsampled corpora,
(2) Stability of word2vec (w2v) embeddings across five downsampled corpora, and
(3) Stability of word2vec embeddings using five random seeds on one downsampled corpus. 
For the Bible, we only show the third case, since it is too small for downsampling.

\begin{figure}[!tb]
\centering
\begin{subfigure}[b]{0.48\textwidth}
   \includegraphics[width=\textwidth]{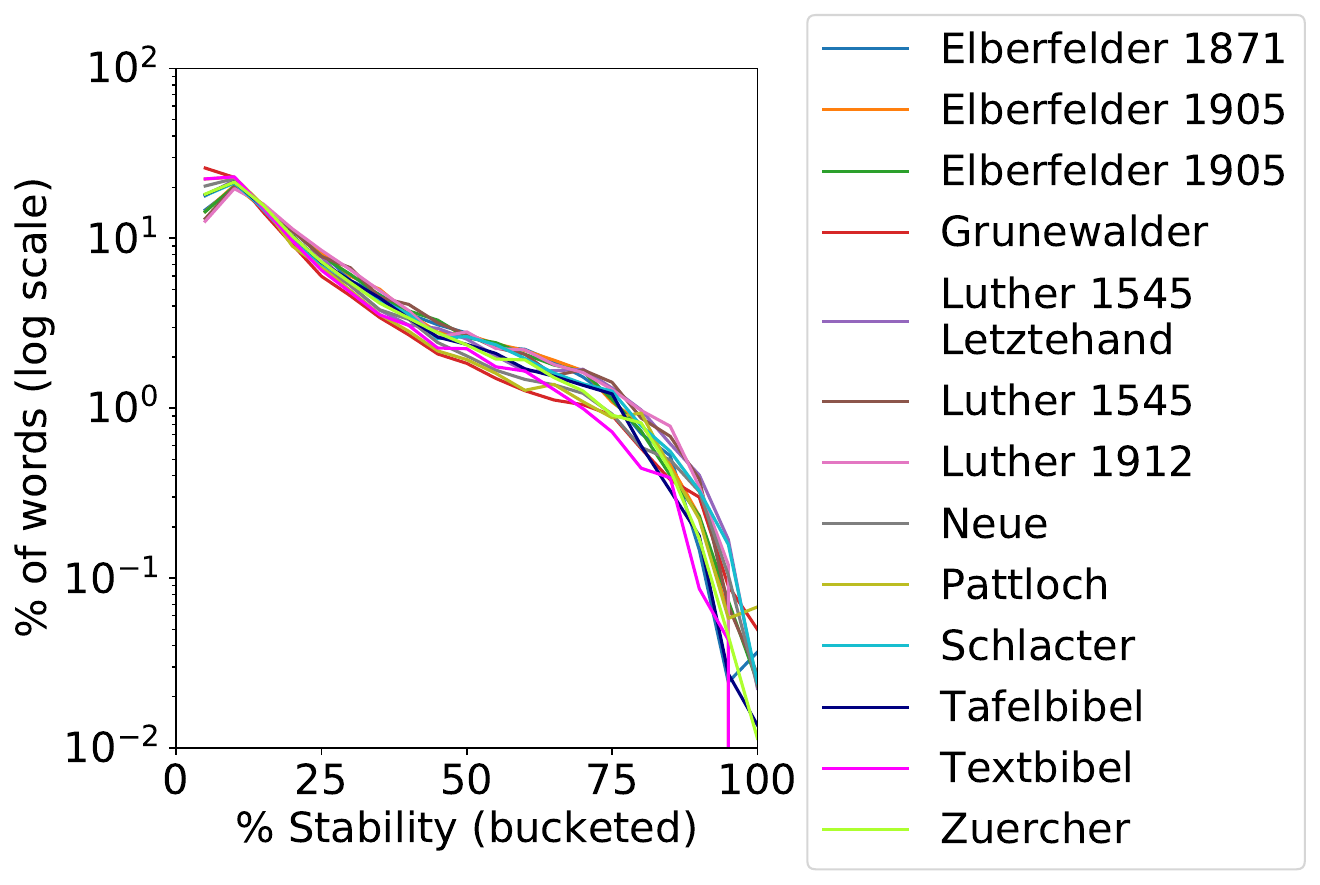}
   \caption{German}
   \label{fig:German} 
\end{subfigure}

\begin{subfigure}[b]{0.48\textwidth}
   \includegraphics[width=\textwidth]{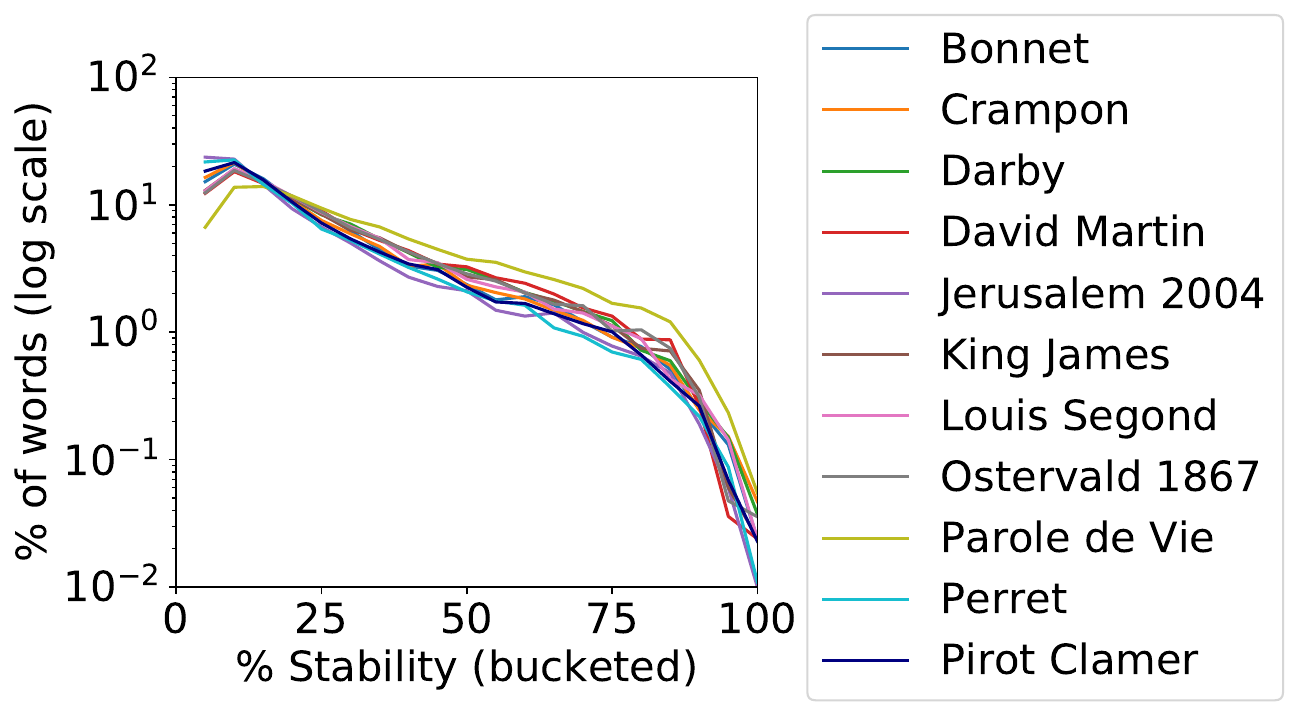}
   \caption{French}
   \label{fig:French} 
\end{subfigure}

\caption{Percentage of words that occur in each stability bucket for different Bible translations.}
\label{fig:sameLanguage}
\end{figure}

Each downsampled corpus is 100,000 sentences, and words that occur with a frequency less than five are ignored. Previous work \cite{pierrejean-tanguy-2018-towards} has indicated that words that appear this infrequently will be very unstable.
We use standard parameters for both embedding algorithms.\footnote{For GloVe \cite{pennington2014glove}, we use 100 iterations, 300 dimensions, a window size of 5, and a minimum word count of 5; these parameters led to good performance in \citeauthor{Wendlandt18Surprising} \shortcite{Wendlandt18Surprising}.
For word2vec \cite{mikolov2013distributed}, we use 300 dimensions, a window size of 5, and a minimum word count of 5.}
For each embedding, we calculate the ten nearest neighbors of every word using FAISS\footnote{We use exact, not approximate, search.} \cite{JDH17}. Finally, for each language, we calculate the stability for every word in that language across all five embedding spaces. Experimental runtimes are listed in Table~\ref{tab:runtimes}.

Figure~\ref{fig:polyglot_methods} shows bucketed stability for both Wikipedia and the Bible.
Most languages have the same overall trend: a large number of relatively unstable word embeddings, then a fairly flat distribution between 25\% and 75\%, and a sharp drop at high stability.
This indicates that the conclusions from prior work on English apply to other languages as well.
In particular, it means that any work that uses embeddings to study a language should train multiple embedding spaces to ensure robust findings.

Some languages have substantially more stable embeddings than others.
Comparing GloVe downsamples on Wikipedia, Vietnamese has the most stable embeddings (avg. 2.46\%), while Korean has the least stable embeddings (avg. 0.58\%). The plot for Vietnamese has a different trend than many of the other plots in Figure~\ref{fig:polyglot_methods}. Vietnamese is the only Austro-Asiatic language in our dataset, so there could be multiple distinctives that are related to it exhibiting different patterns than the other languages.

Finally, varying the training algorithm has a smaller impact than changing the dataset.
Keeping the dataset fixed (Wikipedia) and varying the algorithm, we see similar trends.
Keeping the algorithm fixed (w2v random seeds) and varying the dataset, we often see substantial shifts.
This means that in order to compare languages we need to carefully control for the content of the corpus (which the Bible data allows us to do).
While the Bible is too small to support downsampling, these results on Wikipedia suggest that experiments varying the random seed lead to similar variations to experiments varying the data sample.

To confirm this finding, we consider two languages with multiple Bible translations: German and French. We average stability across five word2vec embeddings using five random seeds on one downsampled corpus. The downsampled corpus is 100,000 sentences, randomly sampled.
Figure~\ref{fig:sameLanguage} shows the stability patterns for each.
The results are very consistent, indicating that variations in translator behavior do not impact stability the way shifting from one corpus to another does.
The largest shift is for the French Parole de Vie translation (top line in yellow in Figure~\ref{fig:French}), which intentionally uses simpler, everyday language.
For further experiments on languages with multiple Bible translation, we choose the Bible translation with the highest average stability.

It is difficult to infer more from these figures alone.
In the next section, we use regression modeling to identify patterns in the results.
Based on the observations above, we use results from GloVe across five downsampled corpora for Wikipedia, and results across five random seeds for the Bible.

\section{Regression Modeling}\label{sec:regression}
We now explore linguistic factors that correlate with stability.
To draw conclusions about specific linguistic features, we use a ridge regression model \cite{hoerl1970ridge}\footnote{Run using the Python package {\tt sklearn.linear. model.Ridge} \cite{scikit-learn} with default parameters except $\alpha=10$.} to predict the average stability of all words in a language given features reflecting language properties. Regression models have previously been used to measure the impact of individual features \cite{singh2016quantifying}. Ridge regression regularizes the magnitude of the model weights, producing a more interpretable model than non-regularized linear regression. We experiment with different regularization strengths and use the best-performing value ($\alpha=10$).\footnote{We run leave-one-language-out cross-validation, described in Section \ref{sec:evaluation}, using the $\alpha$ values of  0.0001, 0.001, 0.01, 0.1, 1, 10, 100, and 1000, choosing the $\alpha$ value with the lowest average absolute error.}
We choose to use a linear model here because of its interpretability. While more complicated models might yield additional insight, we show that there are interesting connections to be drawn from a linear model. 

\subsection{Model Input and Output}
Our model takes linguistic features of a language as input and predicts stability as output. Since WALS properties are categorical, we turn each property into a set of binary features. If a particular language does not have a known value for a given property, then all of these features are marked zero.

In order to draw out important correlations between linguistic features and stability, we filter the languages and WALS properties that we consider.
We only include languages that have at least 25\% of all WALS properties.
Then, we only consider WALS properties that cover at least 25\% of the filtered languages.
We remove all WALS properties that do not have at least two features that each include at least five languages. 
Note that because all of our input features are binary, all weights are easily comparable.
After this filtering, we end up with 37 languages,\footnote{Bengali, Bulgarian, Cherokee, Comanche, English, Estonian, Finnish, Haitian, Haitian Creole, Hebrew, Hindi, Hmong Njua, Hungarian, Indonesian, Italian, Japanese, Korean, Latin, Latvian, Linda, Lithuanian, Ma'di, Mam, Mandarin, Maybrat, Norwegian, Persian, Pohnpeian, Polish, Portuguese, Russian, Somali, Spanish, Swedish, Thai, Turkish, Ukrainian, Vietnamese} and 97 WALS properties.

We also group highly correlated WALS features. We create the groupings by combining features with a Pearson correlation greater than 0.8. A feature is included in a particular grouping if it correlates highly with any of the features already in the group. Each grouped feature is marked as one if \textit{any} of the included features are marked as one.

For each model, we bootstrap over the input features $1,000$ times, allowing us to calculate standard error for the $R^2$ score and the model weights. Calculating significance for each feature allows us to discard highly variable weights and focus on features that consistently contribute to the regression model, giving us more confidence in the results.

The output of our model is the average stability of a language, which is calculated by averaging together the stability of all of the words in a language.
If a language is present in both corpora, we average the stabilities from the two corpora.

\begin{table*}[!th]
\small
    \centering
    \begin{tabular}{p{0.5cm} p{12.5cm} r}
    \toprule
    Cat. & WALS Attribute & Weight \\
    \midrule
    \parbox{0.5cm}{VC, M} & \parbox{12.5cm}{\textit{Suffixing Grouping:} \\
    \hspace*{0.2cm}$\cdot$Prefixing vs.~Suffixing in Inflectional Morphology: \textbf{Strongly Suffixing}; \\
    \hspace*{0.2cm}$\cdot$Position of Tense-Aspect Affixes: \textbf{Tense-aspect suffixes} \\[-6pt]} & \cellcolor{blue!42}$-0.14\pm0.0$\\
    L & Hand and Arm: \textbf{Different} & \cellcolor{blue!33}$-0.11\pm0.0$\\
    CS & Relativization on Obliques: \textbf{Gap} & \cellcolor{blue!30}$-0.10\pm0.0$\\
    VC & Overlap between Situational \& Epistemic Modal Marking: \textbf{Overlap for both possibility \& necessity} & \cellcolor{blue!27}$-0.09\pm0.0$\\
    NC & Ordinal Numerals: \textbf{First, second, three-th} & \cellcolor{blue!24}$-0.08\pm0.0$\\
    NC & Comitatives and Instrumentals: \textbf{Differentiation} & \cellcolor{blue!24}$-0.08\pm0.0$\\
    P & Rhythm Types: \textbf{Trochaic} & \cellcolor{blue!24}$-0.08\pm0.0$\\
    WO & Order of Adjective and Noun: \textbf{Adjective-Noun} & \cellcolor{blue!21}$-0.07\pm0.0$\\
    WO & Order of Adposition and Noun Phrase: \textbf{Postpositions} & \cellcolor{blue!21}$-0.07\pm0.0$\\
    \midrule
    NC & \parbox{12.5cm}{\textit{No Gender Grouping:} \\
    \hspace*{0.2cm}$\cdot$ Systems of Gender Assignment: \textbf{No gender}; \\
    \hspace*{0.2cm}$\cdot$ Sex-based and Non-sex-based Gender Systems: \textbf{No gender}; \\
    \hspace*{0.2cm}$\cdot$ Gender Distinctions in Independent Personal Pronouns: \textbf{No gender distinctions}; \\
    \hspace*{0.2cm}$\cdot$ Number of Genders: \textbf{None} \\[-6pt]} & \cellcolor{red!15}$0.05\pm0.0$\\
    P & Voicing and Gaps in Plosive Systems: \textbf{Other} & \cellcolor{red!18}$0.06\pm0.0$\\
    M & Prefixing vs. Suffixing in Inflectional Morphology: \textbf{Little affixation} & \cellcolor{red!18}$0.06\pm0.0$\\
    CS & `Want' Complement Subjects: \textbf{Subject is expressed overtly} & \cellcolor{red!18}$0.06\pm0.0$\\
    VC & The Morphological Imperative: \textbf{No second-person imperatives} & \cellcolor{red!18}$0.06\pm0.0$\\
    CS & Purpose Clauses: \textbf{Balanced} & \cellcolor{red!18}$0.06\pm0.0$\\[1pt]
    WO & \parbox{12.5cm}{\textit{Prepositions Grouping:} \\
    \hspace*{0.2cm}$\cdot$Order of Adposition and Noun Phrase: \textbf{Prepositions}; \\
    \hspace*{0.2cm}$\cdot$Relationship between the Order of Object and Verb and the Order of Adposition and Noun Phrase: \hspace*{0.25cm}\textbf{VO and Prepositions} \\[-6pt]} & \cellcolor{red!18}$0.06\pm0.0$\\
    WO & Order of Demonstrative and Noun: \textbf{Noun-Demonstrative} & \cellcolor{red!21}$0.07\pm0.0$\\
    NC & Position of Case Affixes: \textbf{No case affixes or adpositional clitics} & \cellcolor{red!33}$0.11\pm0.0$\\
    \bottomrule
    \end{tabular}
    \caption{Weights with the highest magnitude in the regression model. Negative weights correspond with low stability, and positive weights correspond with high stability.}
    \label{tab:allLanguages_weights}
\end{table*}

\subsection{Evaluation} \label{sec:evaluation}
We evaluate our model in two ways. First, we measure goodness of fit using the coefficient of determination $R^2$.\footnote{Measured using the Python package\\ {\tt sklearn.linear\_model.Ridge.score}.}
This measures how much variance in the dependent variable $y$ (average stability) is captured by the independent variables $x$ (WALS properties). A model that always predicts the expected value of $y$, regardless of the input features, will have an $R^2$ score of 0. The highest possible $R^2$ score is 1, and $R^2$ can be negative. Second, in addition to the $R^2$ score, we run leave-one-out cross-validation across all languages, and report absolute error on the left-out language. We compare this to a baseline of choosing the average stability over all training languages.

We use the individual feature weights to measure how much a particular feature contributes to the overall model.
When reporting weights, we train the model using all 37 languages. 
Because we are primarily using regression modeling to learn associations between certain features and stability, no test data are necessary. The emphasis is on the model itself and the feature weights it learns, not on the model's performance on a task.

\section{Results and Discussion}\label{sec:results}

Our regression model has a high $R^2$ score of $0.96\pm0.00$, indicating that the model fits the data well. 
Significant weights with the highest magnitude are shown in Table \ref{tab:allLanguages_weights}. Running leave-one-out cross-validation across all languages, we get an average absolute error of $0.62\pm 0.53$.
\footnote{Cross-validation has an average $R^2$ score of $0.92$ on the training data.} 
For comparison, using the average stability gives an average absolute error of $0.86\pm 0.55$. (A two sample t-test comparison gives a p-value of $0.060$.)

\begin{table}[!t]
    \centering
    \scalebox{0.92}{
    \begin{tabular}{lrr}
    \toprule
    WALS Category & Num. & Avg. \\
    & Features & Magnitude \\
    \midrule
    Simple Clauses (SC) & 30 & 0.019 \\
    Nominal Syntax (NS) & 2 & 0.021 \\
    Other (O) & 2 & 0.023 \\
    Complex Sentences (CS) & 11 & 0.028 \\
    Morphology (M) & 18 & 0.031 \\
    Word Order (WO) & 32 & 0.031 \\
    Phonology (P) & 21 & 0.032 \\
    Nominal Categories (NC) & 40 & 0.036 \\
    Verbal Categories (VC) & 27 & 0.036 \\
    Lexicon (L) & 6 & 0.039 \\
    \bottomrule
    \end{tabular}
    }
    \caption{Number of binary features and average magnitude of weights in the regression model for different WALS categories. Grouped features are included in each category that they cover.}
    \label{tab:wals_categories}
\end{table}

\begin{figure}[!t]
\centering
\begin{subfigure}[b]{0.45\textwidth}
   \includegraphics[width=\textwidth]{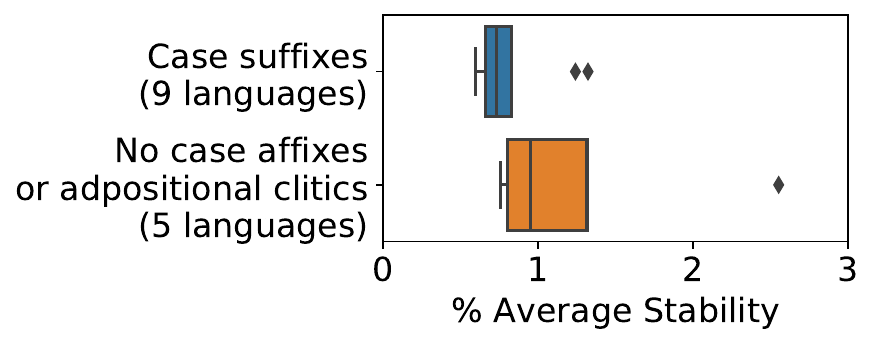}
   \caption{Position of Case Affixes}
   \label{fig:51a} 
\end{subfigure}
\vspace{0.2cm}

\begin{subfigure}[b]{0.45\textwidth}
   \includegraphics[width=\textwidth]{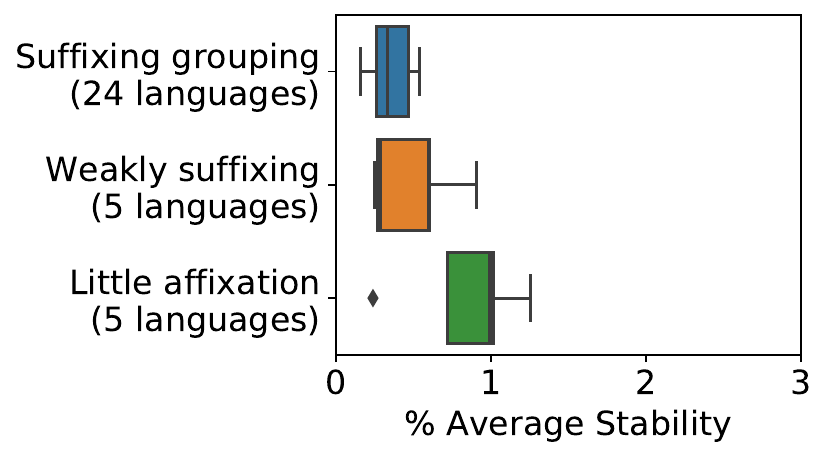}
   \caption{Prefixing vs. Suffixing in Inflectional Morphology}
   \label{fig:26} 
\end{subfigure}
\vspace{0.2cm}

\begin{subfigure}[b]{0.45\textwidth}
   \includegraphics[width=\textwidth]{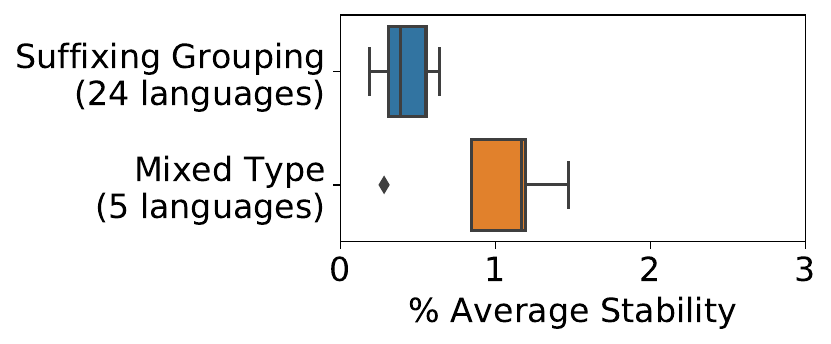}
   \caption{Position of Tense-Aspect Affixes}
   \label{fig:69} 
\end{subfigure}

\caption{Affixing properties compared using box-and-whisker plots.}
\label{fig:boxplots}
\end{figure}

Table \ref{tab:wals_categories} breaks down the regression results by broad WALS category, listing both the number of binary features per category, as well as the average magnitude of weights for features in that category.
The two most important groups of features are Nominal Categories and Verbal Categories. Both of these categories have a large number of features and a high average magnitude. While the Lexicon category has a high average magnitude, it contains very few features. To further explore these results, we highlight a few WALS property in more detail.

\mysubsection{Suffixes and prefixes.}
Table \ref{tab:allLanguages_weights} shows that three of the top features are related to affixes (suffixes and prefixes). Specifically, three main properties deal with affixes: Position of Case Affixes \cite{wals-51}, Prefixing vs. Suffixing in Inflectional Morphology \cite{wals-26}, and Position of
Tense-Aspect Affixes \cite{wals-69}. Distributions of these features in the 37 languages used for the regression model are shown in Figure \ref{fig:boxplots} (categories with fewer than five languages are not shown).

\begin{figure}[!t]
\centering
\begin{subfigure}[b]{0.45\textwidth}
   \includegraphics[width=\textwidth]{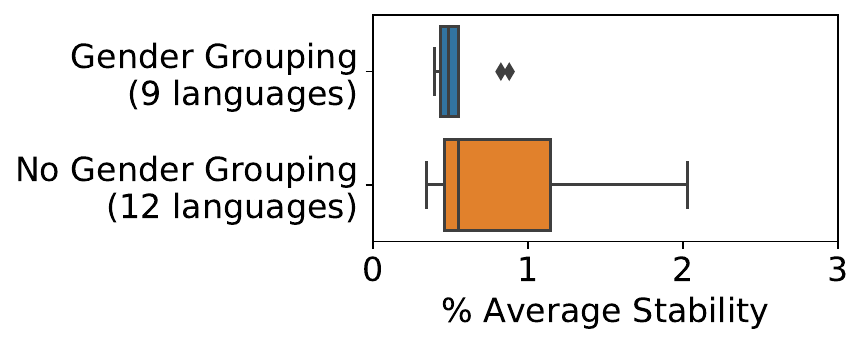}
   \caption{Systems of Gender Assignment; Sex-based and Non-sex-based Gender Systems (\textit{Gender Grouping:} No gender; Sex-based) }
   \label{fig:gender1} 
\end{subfigure}
\vspace{0.2cm}

\begin{subfigure}[b]{0.45\textwidth}
   \includegraphics[width=\textwidth]{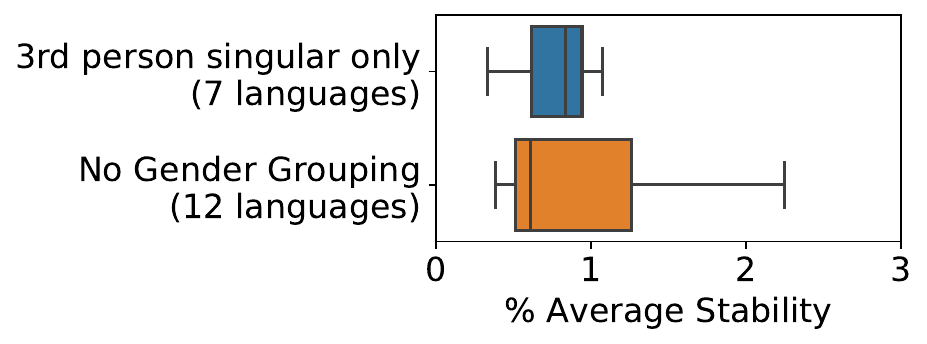}
   \caption{Gender Distinctions in Independent Personal Pronouns}
   \label{fig:gender2} 
\end{subfigure}
\vspace{0.2cm}

\begin{subfigure}[b]{0.45\textwidth}
   \includegraphics[width=\textwidth]{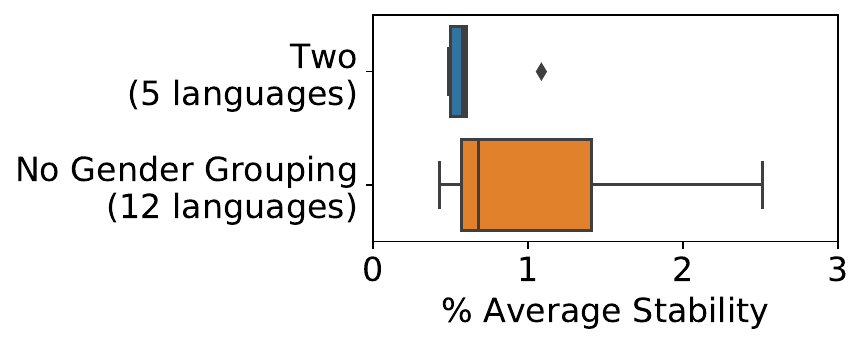}
   \caption{Number of Genders}
   \label{fig:gender3} 
\end{subfigure}

\caption{Gender properties compared using box-and-whisker plots.
Note, the 12 languages with ``No Gender Grouping'' are not the same across the three plots.}
\label{fig:boxplots2}
\end{figure}

For all three of these properties, more affixing is associated with lower stability. When considering word embeddings, this result makes intuitive sense. Affixes cause there to be many different word variations (e.g., \textit{walk}, \textit{walked}, \textit{walking}, \textit{walker}), which may not be handled consistently by the embedding algorithm, leading to lower average stability.

\mysubsection{Gendered Languages.}
Table \ref{tab:allLanguages_weights} also highlights a grouping of WALS properties related to whether a language is gendered or not. Four WALS properties are relevant to this: Systems of Gender Assignment \cite{wals-32}, Sex-based and Non-sex-based Gender Systems \cite{wals-31}, Gender Distinctions in Independent Personal Pronouns \cite{wals-44}, and Number of Genders \cite{wals-30}. In general, a language is considered to have a gender system if different parts-of-speech are required to agree in gender (as opposed to simply having gendered nouns). Distributions of these features are shown in Figure \ref{fig:boxplots2}.

For all of these properties, languages with no gender system tend to have higher average stability. Again, this result makes sense in the context of word embeddings. Languages with gender systems will have more word forms (e.g., both male and female word forms), which may not be handled consistently by the embedding algorithm.

\section{Conclusion}

In this paper, we considered how stability varies across different languages. This work is important because algorithms such as GloVe and word2vec continue to be effective methods in a wide variety of scenarios \cite{Arora2020ContextualEW}, particularly the computational humanities and languages where large corpora are not available. We studied the relationship between linguistic properties and stability, something that has been previously understudied.
We drew out several aspects of this relationship, including that languages with more affixing tend to have less stable embeddings, and languages with no gender systems tend to have more stable embeddings. These insights can be used in future work to inform the design of embeddings in many languages. For example, this work suggests that future embedding space designs need to take into account gendered words and morphologically rich words with affixes.

\section{Acknowledgements}

This material is based in part upon work supported by the National Science Foundation (grant \#1815291) and by the John Templeton Foundation (grant \#61156). Any opinions, findings, and conclusions or recommendations expressed in this material are those of the author and do not necessarily reflect the views of the National Science Foundation or John Templeton Foundation. 

\bibliography{output_paper}
\bibliographystyle{emnlp21/acl_natbib}

\end{document}